\documentclass[A4paper, 11 pt, conference]{ieeeconf}  
\IEEEoverridecommandlockouts       
\usepackage{geometry}
\usepackage{amsmath}
\usepackage{amssymb}
\usepackage{amsfonts}
\usepackage{algorithm}
\usepackage{algpseudocode}
\usepackage{graphicx}
\usepackage{epsfig}
\usepackage{tensor}
\usepackage{color}
\usepackage{float}
\usepackage{booktabs}
\usepackage{wasysym}
\usepackage{array}
\usepackage[dvipsnames]{xcolor}

\newcommand*\diff{\mathop{}\!\mathrm{d}}
\DeclareMathOperator{\sgn}{sgn}

\graphicspath{{figures/}}
\DeclareGraphicsExtensions{.eps}

\begin{document}

\title{\LARGE \bf
Can a Tesla Turbine be Utilised as a Non-Magnetic Actuator for MRI-Guided Robotic Interventions?}

\author{David Navarro-Alarcon, Luiza Labazanova, Man Kiu Chow, Kwun Wang Ng and Derek Kwok
\thanks{D. Navarro-Alarcon, L. Labazanova and M. Chow are with The Hong Kong Polytechnic University, KLN, Hong Kong.}%
\thanks{K. W. Ng is with the Chinese University of Hong Kong, Shatin, Hong Kong.}
\thanks{D. Kwok is with Time Medical Ltd, Shatin, Hong Kong.}
}

\maketitle 
\thispagestyle{empty}

\begin{abstract}
This paper introduces a new type of non-magnetic actuator for MRI interventions. 
Ultrasonic and piezoelectric motors are one the most commonly used actuators in MRI applications.
However, most of these actuators are only MRI-safe, which means they cannot be operated while imaging as they cause significant visual artifacts. 
To cope with this issue, we developed a new pneumatic rotary servo-motor (based on the Tesla turbine) that can be effectively used during continuous MR imaging. 
We thoroughly tested the performance and magnetic properties of our MRI-compatible actuator with several experiments, both inside and outside an MRI scanner.
The reported results confirm the feasibility to use this motor for MRI-guided robotic interventions.
\end{abstract}

\section{Introduction}
Magnetic resonance imaging (MRI) is a medical imaging technique that creates detailed anatomical pictures of internal structures of the body.
A common application of MRI scanners is for guiding of needles during biopsy interventions (where sample tissues are extracted from an area of interest). 
Compared to other medical imaging modalities, MRI offers several advantages when performing interventional procedures, amongst them are: MRI does not expose the patient to harmful radiation (e.g. as with x-ray machines), it offers excellent soft tissue contrast for locating lesions, it produces detailed spatial information of tissues in 3D, to name a few \cite{lee2010breast}. 
With the aim of improving the precision and dexterity of these interventions, robotic technologies have been recently introduced to the MRI room.
However, since the operation principle of MRI is based on strong magnetic fields, traditional actuation technologies (e.g. standard electric servo-motors) cannot be used to drive the motion of an interventional robot; conventional electronic and signal transmission systems used in the actuator's controls severely affect the imaging process.
Furthermore, the high magnetic field also imposes strict requirements to the types of materials that can be used in an actuator operating nearby the magnetic bore; ferromagnetic metals should be avoided as they cause image artifacts.

To understand the extent to which devices (including actuators and sensors) can operate within an MRI bore, we must first introduce two definitions.
A device is considered to be \emph{MRI-safe} if it does not present any potential risk to patients or others, but may introduce noise into the images. 
A device is considered to be \emph{MRI-compatible} if it is MRI-safe and does not significantly affect the imaging quality and the operation of the scanner \cite{tsekos2007magnetic} (clearly, MRI-safe actuators may only be applicable to procedures that can be conducted/guided with off-line scans).
In the past two decades, many researchers that have developed both MRI-safe and MRI-compatible actuation systems for various types of interventions \cite{Journals:Elhawary2008}. 
Based on their operation principle, these actuators can be classified into three main categories: (1) piezoelectric/ultrasonic motors, (2) hydraulic actuators, and (3) pneumatic actuators.

Piezoelectric/ultrasonic motors are frequently found in the MRI robotics literature.
These types of actuators produce motion based on the non-magnetic piezoelectric effect, that allows them to safely be brought inside the bore. 
Some examples of these systems are given in \cite{Hata2005}, where a thermo-therapy robot for liver tumors is proposed, and \cite{Flueckiger2005}, where a haptic interface controlled by an ultrasonic motor is developed). 
In order to generate motion, this type of motors are activated with high frequently signals that may affect the signal-to-noise ratio (SNR) and cause severe degradation to the images. 
Many authors stress that these actuators should not be operated while performing MRI scans, see \cite{Journals:Masamune1995,Journals:Mozer2009}.
Piezoelectric/ultrasonic are only MRI-safe.

Hydraulic actuators use controllable liquid flows to generate driving forces.
Examples of this technique are given in \cite{Moser2003}, which describes a master-slave interface for studying motor control, and \cite{Raoufi2008} which presents robot for a neurosurgery application. 
Hydraulic actuation provides high driving forces (thanks to the incompressibility of liquids), can be accurately controlled, and is MRI-compatible. 
However, hydraulic devices and installations tend to create bulky setups and, most importantly, suffer from fluid leakage that might contaminate the MRI room. 

Pneumatic actuation is a cleaner and easier to maintain option that relies on compressed air to power a system. 
Pneumatics is MRI-compatible, and according to some studies (see e.g. \cite{Journals:Fischer2008}) it presents the best performance for continuous imaging applications.
An early application is given \cite{Hempel2003}, where a robotic needle driver for radiological interventions is reported.
The authors in \cite{stoianovici2007new,Muntener2008} developed a pneumatic stepping rotary motor (based on a planetary mechanism) that has an angular step of 3.33 degrees and output torques of several hundred of N-mm. 
Another stepping motor was developed in \cite{chen2014mr}; this system uses two air cylinders to form a crank-link mechanism that outputs rotary motions. 
Recently, the authors in \cite{groenhuis2016laser} developed new kind of stepping robot (using laser cutting technology) for needle positioning applications.
These types of stepping systems are characterised by moving through discrete motions.
Note that resonances might arise when rotating around the system's natural frequency; accelerations are also difficult to control.

To cope with the above-mentioned issues, we have developed a new MRI-compatible pneumatic actuator that allows to effectively generate continuous smooth rotary motions.
The basic mechanical structure of this new system is based on the Tesla turbine.
To guarantee that it can operate under continuous MR imaging, it is fabricated using 3D printing technology, and its angular position is measured using fibre optics.
Experiments are conducted to verify the motor's MRI-compatibility and performance.

The rest of this paper is organised as follows: Section \ref{sec:methods} describes the design of the new actuator; Section \ref{sec:results} presents the conducted experimental study; Section \ref{sec:conclusions} gives final conclusions.

\section{Methods}\label{sec:methods}
\subsection{Mechanics of the Tesla Motor}
\begin{figure}[!ht]
	\centering
	\includegraphics[width=\columnwidth]{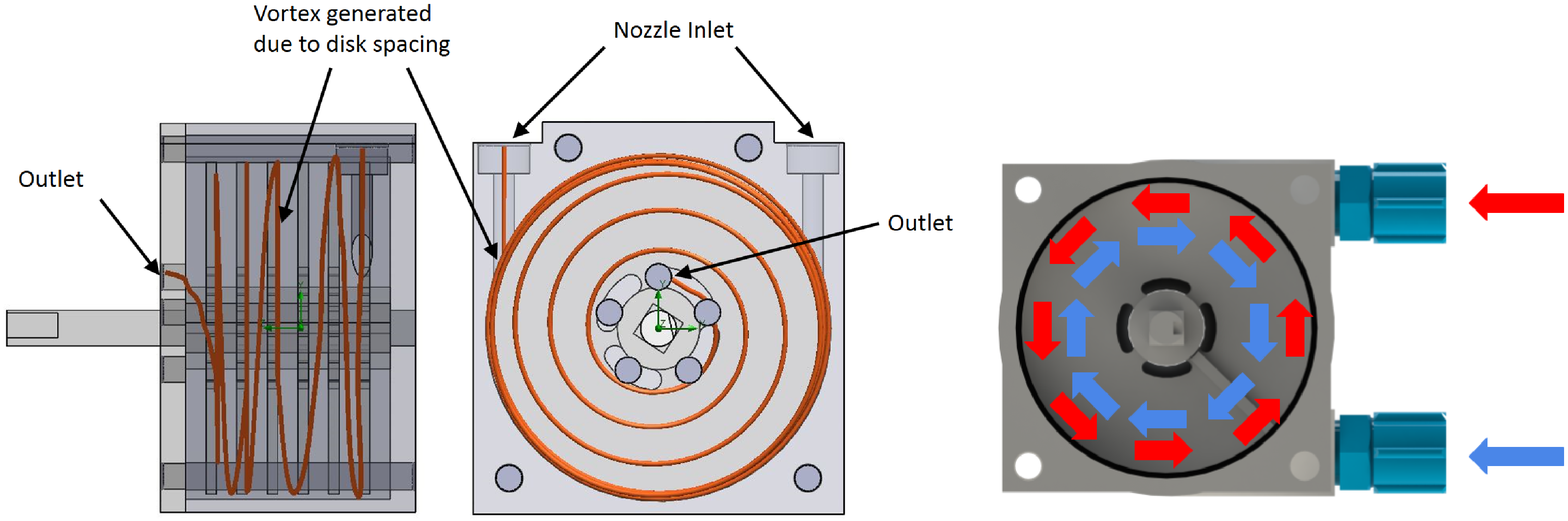}
	\includegraphics[width=\columnwidth]{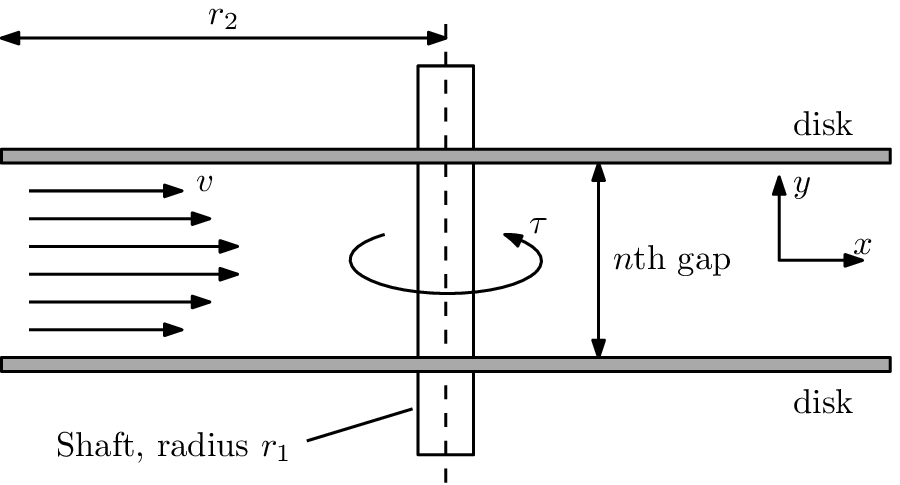}
	\caption{Working principle of a pneumatic actuator}
	\label{Tesla Principle}
\end{figure}

In this section we present the development of the pneumatically actuated motor, whose motion principle is based on the Tesla turbine, which was developed by Nicola Tesla in 1905 as a hydroelectric power generator \cite{vsarboh2010patents} (along this paper, we shall refer to the application of this system as the \emph{Tesla motor}).
The proposed motor is a blade-less turbine that operates based on the boundary layer effect of the driving fluid. 
It consists of a set of several smooth disks separated by a small gap, and that are fixed together to central rotating shaft. 
Each disk is provided with four exhaust holes that are placed near to the disk's centre. 
When compressed air flows into the motor through the inlet nozzle, it spirals around the shaft and moves towards the exhaust ports creating a vortex.
The fluid vortex induces a drag force over the disks' surface, that results in rotational motion of the motor shaft.
The proposed motor is equipped with two independent inlet ports that enable bi-directional rotations (Figure \ref{Tesla Principle} shows the schematic drawing).

We mathematically model the motor's driving torque by analysing its fluid-structure interactions.
Consider the velocity profile $v$ of the fluid (assumed to be parabolic) that moves along the tangential direction relative to the disks' surface.
This tangential flow between two consecutive disks generates a shear stress that is proportional to the fluid's friction coefficient $\mu$.
The total driving torque $\tau$ that is generated by a motor with $n$ disk gaps satisfies the following relation \cite{guha2013fluid}:
\begin{equation}
	\tau = n \mu \int_{r_1}^{r_2} \frac{\partial v}{\partial y} \,dr
	\label{Eq:torque_fluid}
\end{equation}
where $r_1$ and $r_2$ are the radius of the outlet at the centre of the disk and the outer radius of the disk, respectively, and $y$ denotes the normal direction to the disk's surface (see Figure \ref{Tesla Principle} bottom).
The velocity $v$ is proportional to the pressure of the driving compressed air system, and its monotonic (yet nonlinear) relation can be used for controlling the motion of the motor (and it will be shown later).

\subsection{Motor Prototyping}
Our objective is to develop an MRI-compatible actuator that can operate within the magnetic bore.
Therefore, in the design and fabrication of the motor it is important to select non-magnetic materials for its components. 
The motor's structure was 3D printed with PLA (polylactic acid) using a standard 3-axis printer in our laboratory; only the disks and spacers were printed by resin using SLA (stereo-lithography) printing. 
The disks were fabricated with this latter method since it produces a much smoother surface that helps to create stable airflow for driving the motor. 
The disks printed with PLA have a much rougher surface, even when using the smallest layer achievable by our 3D printer ($0.06$ mm in our case); if this method used to build the motor, the rough finishing causes unsteady airflow that affects the motor's performance. 
It is important to remark that since the disks are fabricated with plastics, they have less rigidity and strength compared to metal-based disks (they may be susceptible to bending when the compressed air first strikes the disk next the inlet port).
For our MRI robotics application, we use non-magnetic metals (viz. brass and aluminium) only for a few essential support and driving components. 
By using this type of metals, we aim to reduce the overall magnetic susceptibility of the actuator.

\begin{figure}[!ht]
	\centering
	\includegraphics[width=.55\columnwidth]{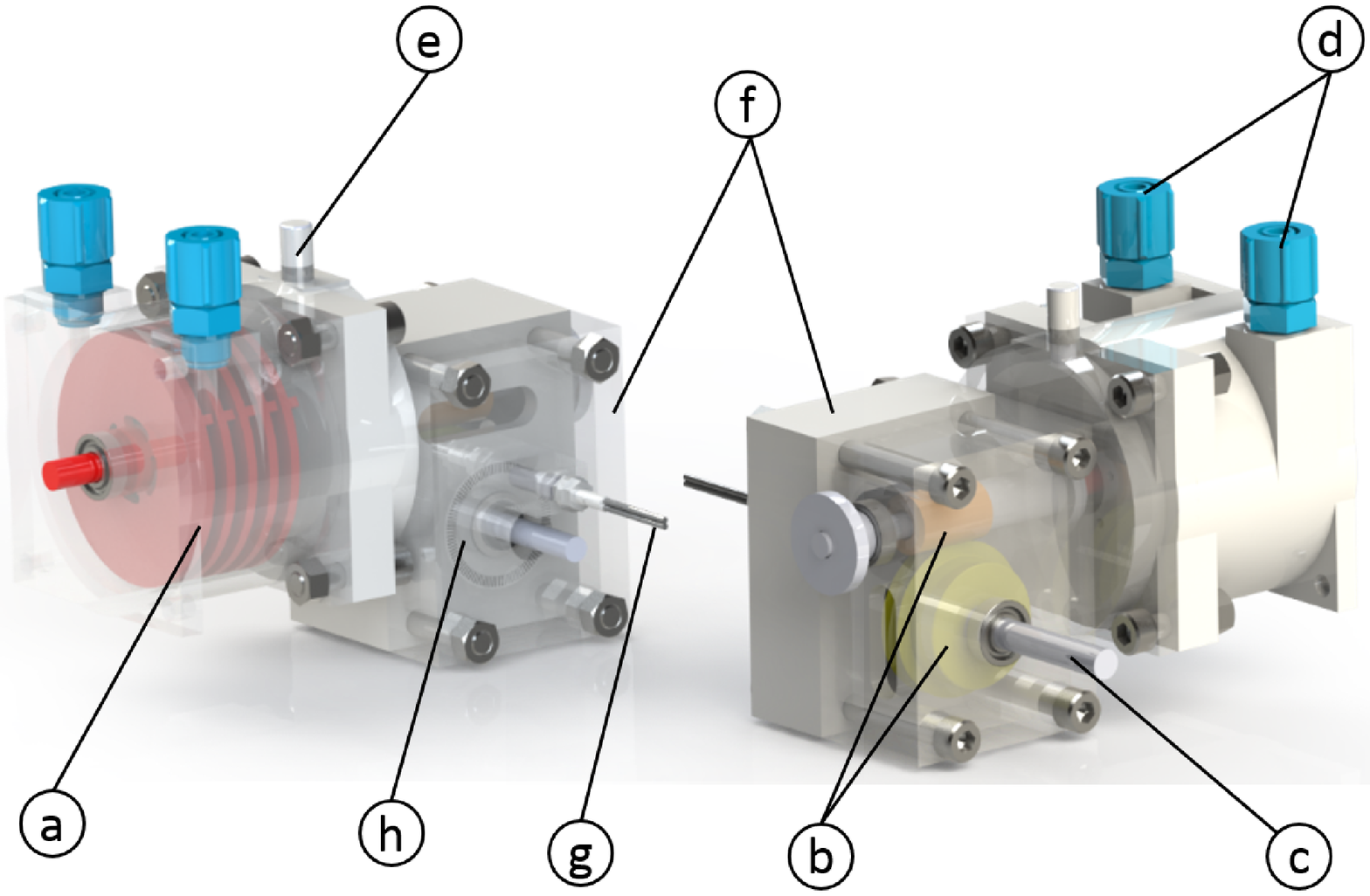}
	\includegraphics[width=.43\columnwidth]{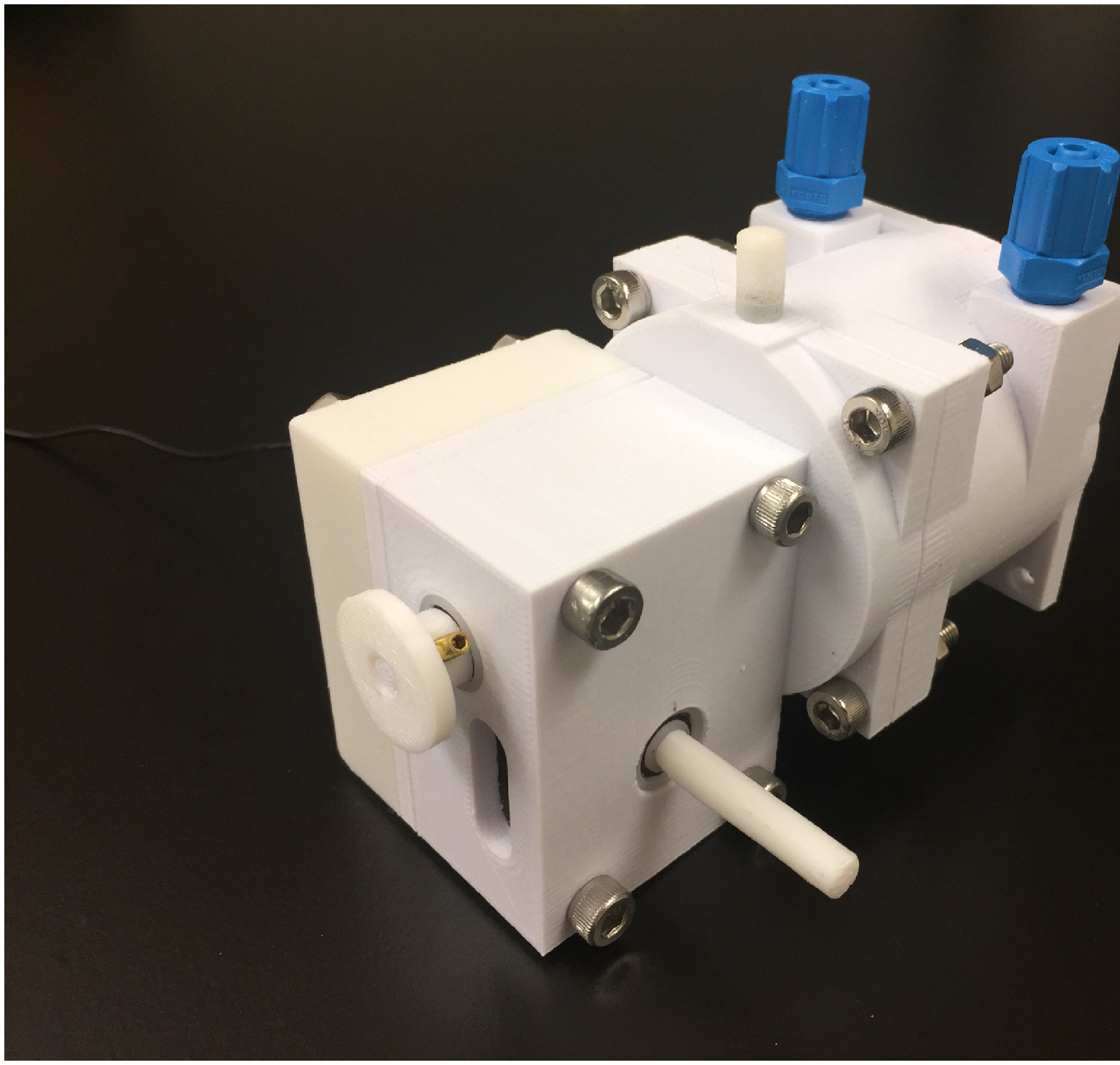}
	\caption{(Left) Schematic drawing of Tesla motor: (a) disks assembly, (b) worm gears, (c) output shaft, (d) inlets ports, (e) silencer, (f) optical fibre, (g) rotary encoder (h) encoder disc. (Right) Prototype of Tesla motor}
	\label{TM Drawing}
\end{figure}

The basic design of a Tesla motor is characterised for generating a high rotational speed with a low driving torque. 
In laboratory tests, sensor feedback shows that the developed motor can achieve a rotational speed of around 13000 RPM when driven by compressed air of 4 Bar. 
To use this motor in a robotic mechanism, we must first modify its speed and torque properties. 
For that, we developed an custom-made MRI-compatible gearbox with a 1:60 gear reduction (some of the components of this gearbox are fabricated with brass and not plastics so as to improve its strength). 
The gearbox is built with worm gears for the following reasons: (1) it provides high gear ratio that effectively increases torque and reduces the output speed; (2) it provides a self-locking feature (i.e. it is non back-drivable) that improves stiffness and prevents undesired motions when disturbances arise; (3) it has a compact structure with fewer components than e.g. planetary gears with the same reduction ratio. 
Figure \ref{TM Drawing} shows the fabricated prototype with its different components.

\begin{figure}[!ht]
	\centering
	\includegraphics[width=\columnwidth]{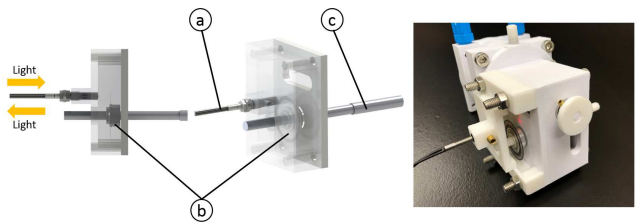}
	\caption{(Left) Details of the proposed rotary encoder, where: (a) optical fibre; (b) encoder disk; (c) output shaft. (Right) The developed prototype for the MRI-compatible sensor.}
	\label{Light encoder}
\end{figure}

The use of electrical signals in a sensor operating inside the magnetic bore can potentially create noise in the image when performing scans.
Therefore, we developed a custom-made rotary encoder that uses optical fibres (from Keynece\textsuperscript{\textregistered}) to measure/estimate the angular position and velocity of the motor.
Since the rotational speed of the turbine's shaft is very high at its nominal operation, there might be many missed counted pulses if the rotary encoder is installed on it.
Therefore, the rotary encoder is instead installed at the shaft but of the gearbox, which has a much smaller rotational speed due to its 1:60 gear reduction ratio (e.g. if we use a nominal 4 Bar driving pressure, the gearbox has a speed of around 200 RPM, which is much easier to detect with simple electronic board).
The input-output speed relation of the gearbox is expressed as follows: 
\begin{equation}
				\label{rpm ratio}
				\omega_{out} = \frac{1}{60}\omega_{turbine}
\end{equation}

\subsection{Application to MRI-Guided Interventions}
\begin{figure}[!ht]
    \centering
    \includegraphics[width=0.9\columnwidth]{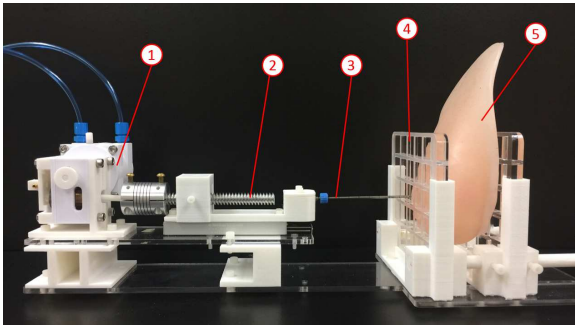}
    \caption{Test set-up for needle insertions: (1) Tesla motor, (2) aluminium power screw, (3) coaxial needle, (4) compression grid, (5) silicon breast phantom.}
    \label{1DOF}
\end{figure}

We developed a test set-up to evaluate the performance and magnetic compatibility of the Tesla motor.
This test system has a 1-DOF linear joint mechanism that uses a non-magnetic slide (Del-Tron\textsuperscript{\textregistered}), an aluminium power screw (Abssac\textsuperscript{\textregistered}), and a nylon nut.
The purpose of the mechanism is to transform the rotary motions of the actuator into controllable linear motions of a biopsy/coaxial needle (hence, simulating in a simple MRI-guided intervention). 
The use of the power screw further reduces the speed of the insertion motions.
Figure \ref{1DOF} shows the developed pneumatically-powered needle insertion mechanism.

Compactness is an important concern for MRI robotics as there is limited space inside the scanner. 
To guarantee that the motor can be used in an interventional robotic system, it is necessary to build it with a moderate size but at the same time it should be able to provide sufficient torque. 
In the developed insertion mechanism, the motor is fabricated with a diameter of ${\diameter}60$ mm and a length of $130$ mm.
Six ${\diameter}55$ mm disks with $2$ mm thick spacers are utilised to generate the driving torque. 
We selected this configuration based on the observed performance of previous test prototypes. 
For a motor with fewer disks, the rotary assembly will have a smaller mass that generates less output torque. 
For a motor with many disks and heavier rotary assembly, the generated (nominal) torque is certainly larger, yet, it requires a higher starting torque to begin the motion.

\subsection{Motion Control System}
\begin{figure}[!ht]
	\centering
	\includegraphics[width=\columnwidth]{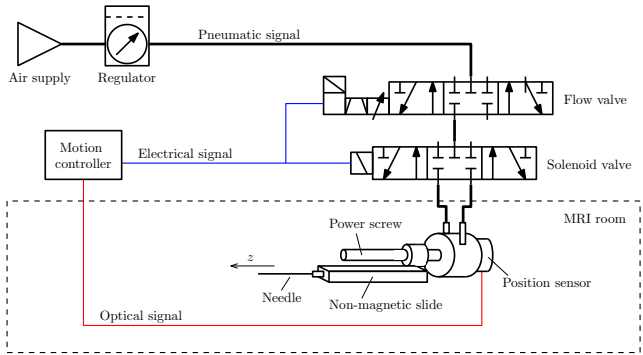}
	\caption{Schematic diagram of the control system}
	\label{Setup_Schematic}
\end{figure}

The control system that commands the motion of the Tesla motor is composed of various parts: a real-time control PC with analogue boards to program the algorithms and output the control actions, a data acquisition system to process the optical sensor feedback, and pneumatic servo valves to regulate the air flow and pressure inside the actuator. 
Figure \ref{Setup_Schematic} shows a schematic diagram of the developed motion controller. 
This diagram shows that to servo-control the motor, only light signals and compressed air need to be passed inside the MRI room; all electrical signals are processed outside the room, therefore, eliminating possible sources of noise.

The Tesla motor is controlled by the action of two pneumatic valves: a flow proportional valve from Festo\textsuperscript{\textregistered} (MPYE-5-1/8-LF-010-B) that regulates the volumetric flow rate that is sent to the motor, and a solenoid valve from SMC\textsuperscript{\textregistered} that directs the flow towards either of the two inlet ports (which determine the directions of rotation).
These pneumatic valves are controlled by an embedded analogue output board from Phidgets\textsuperscript{\textregistered}, that is programmed in a Linux PC with standard C++ language.
Since all the pneumatic devices and controller are placed outside the shielding room, the solenoid valve must be connected to the motor using two $5$ metre-long plastic tubes (which are passed through the scanner's waveguide). 
The motor's rotations are measured by counting the optical pulses from the encoder; these are first processed by the sensor's transceiver, and then acquired into the control PC via the Phidgets board.

In general, to control the motion of pneumatically-driven actuators, one must regulate either the air pressure or the air flow rate. 
For our Tesla motor, the input flow rate $\phi$ is used to indirectly specify the generated output torque \cite{ho2011tesla}.
With this flow-proportional servo-valve, the flow $\phi=\kappa u$ can be accurately set by a commanded analogue voltage $u$, for $\kappa$ as a known parameter of the valve.
We locally approximate the relation (around the nominal operation speed) between $\phi$ and the driving torque $\tau$ with the following expression:
\begin{equation}\label{tesla_torque}
\tau \approx i  \rho h(\phi) = i \rho h(\kappa u)
\end{equation}
where the scalar $i=\pm1$ models the positive/negative direction of rotation (as determined by the solenoid valve), $\rho$ denotes a positive proportionality constant, and $h(\cdot)$ represents a monotonically increasing function of the flow (both $\rho$ and $h(\cdot)$ can be experimentally estimated with standard calibration methods).
From equation \eqref{tesla_torque}, it is clear that the driving torque can be (approximately) adjusted via the analogue voltage that is commanded to the flow valve.
Note that this local model is only valid for a motor under continuous rotations, and does not account for the highly nonlinear static starting torque properties.

The second-order dynamic equations of motion of the Tesla motor are given by:
\begin{equation}\label{tesla_dynamics}
\ddot{q} = \frac{1}{J} (-b\dot{q} - c\sgn(\dot{q}) + \tau_{L} + \tau(u))
\end{equation}
where $J$ denotes the inertia of rotor, $b$ and $c$ are the viscous and the Coulomb friction coefficients, respectively, $\tau_{L} $ represents the torque due to external loads. 
In our system, the motion controller is designed considering the command voltage $u$.
We implemented a standard PID regulator with the following form:
\begin{equation}\label{PIV}
				u = - K_{P}\left( q - q_d \right) + K_{I}\int_{} \left( q - q_d \right)\diff t - K_{D} \dot{q} 
\end{equation}
for $K_p$, $K_i$, and $K_d$ as the proportional, integral and derivative gains, which were experimentally tuned by trial and error, and $q_d$ as the target position. 

\section{Results}\label{sec:results}
We carried out two types of experimental studies to test the new actuator: (1) in-the-lab experiments, and (2) in-bore experiments.
The former were conducted (outside MRI scan room) to evaluate the motor's positioning error, maximum force, and speed-pressure response.
The latter were conducted inside the magnetic bore with the aim of testing the MRI-compatibility, signal-to-noise ratio (SNR), percent integral uniformity (PIU), and homogeneity.

\subsection{In-the-Lab Experiments}
\begin{figure}[!ht]
    \centering
    \includegraphics[width=0.9\columnwidth]{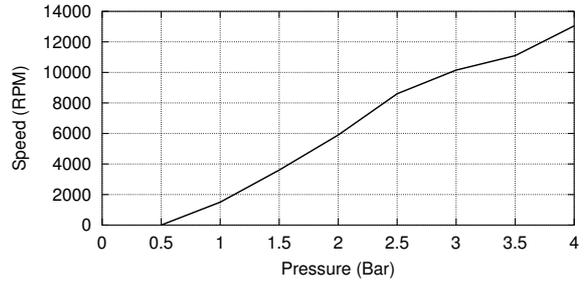}
    \caption{Experimental speed-pressure curve}
    \label{rpms_results}
\end{figure}

We evaluated the resulting rotational speed of the motor under various air pressure inputs.
This type of test is particularly important for our system since some of the 3D printed parts of the rotary disk assembly might melt at high speeds.
Figure \ref{rpms_results} shows the obtained speed-pressure curve. 
From these tests, we observed that the motor shaft starts to deform for speeds higher than 13000 RPM, and that a minimum pressure of around 0.5 Bar must be used in order to overcome the motor's static friction.
It is important to identify these values so as to set both upper and lower pressure limits for achieving a normal operation with the motor.

\begin{figure}[!ht]
	\centering
	\includegraphics[width=\columnwidth]{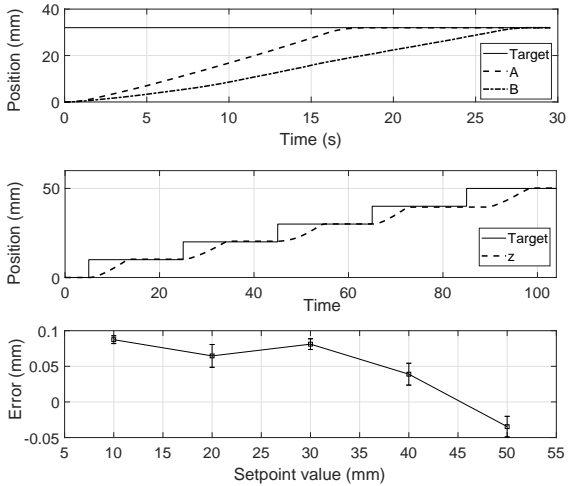}
	\caption{(Top) Positioning curves obtained without (A) and with (B) phantom;. (Middle) Positioning curves for multiple $10$ mm step increments. (Bottom) Position errors for various targets inside the phantom tissue.}
	\label{32mm_results}
\end{figure}

Next, we evaluated the positioning accuracy of the pneumatic motor.
For that, we used the setup shown in Figure \ref{1DOF}, where the task is to introduce the coaxial needle into different targets inside the silicon breast phantom tissue.
The resulting position profiles for a step response of $32$ mm are shown in Figure \ref{32mm_results} top where the labels A and B denote motions with and without the phantom tissue, respectively.
From these results we can see that the joint's motion is slowed down by the opposing forces that result from the interaction with the phantom.
Figure \ref{32mm_results} middle depicts the motor's positioning performance when multiple (smaller) $10$ mm targets are given.
Due to the continuous air injection, the actuator will accumulate considerable kinetic energy for ``large'' target positions (i.e. for greater than $50$ mm).
In this situation, it will exhibit substantial overshoot during the positioning motion (therefore, it is recommended to command motions through small incremental targets or ramp targets).
Figure \ref{32mm_results} bottom depicts the measured position errors $e = q - q_{d}$ for several experiments conducted with the target located inside the phantom tissue.

\begin{figure}[!ht]
    \centering
    \includegraphics[width=0.9\columnwidth]{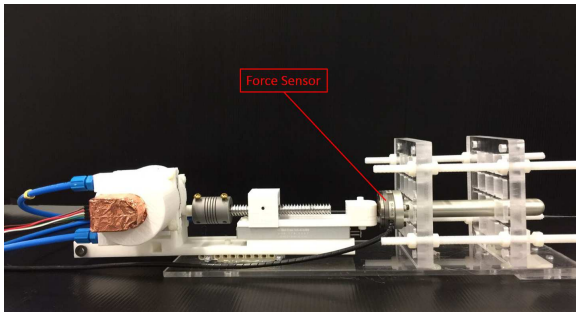}
    \caption{The set-up to test the pressure-force relations.}
    \label{Force sensing}
\end{figure}

The forces that the Tesla motor can generate were also tested.
This experiment is particularly important as we must determine whether the actuator has sufficient power to penetrate human tissues and even tumours.
To conduct these experiments, we use the setup shown in Figure \ref{Force sensing} which has a 6 axis force/moment transducer (ATI Mini40) to measure the generated forces. 
The obtained pressure-force relations are given in Table \ref{force}; these data shows that the coupling of the rotary actuator with the linear power screw can generate forces larger that $10$ N.
It has been reported in previous studies (see e.g. \cite{chun2013experimental,kobayashi2012enhanced}) that an insertion force of around $2$ N is required to introduce a biopsy needle into animal and human breast tissues.
In \cite{kokes2009towards}, it is reported that a maximum $4$ N axial force is required to penetrate a tumour.
These data suggests that the developed Tesla motor can provide enough force to perform a biopsy procedure.

\begin{table}
	\centering
	\caption{The measured pressure-force relations.}
	\label{force} 
	\begin{tabular}{p{3cm}p{3cm}p{4cm}}
					\hline
					\hline
					Pressure (Bar) & Force (N)\\
					\hline
					1.5 & 11.49 \\
					2.0 & 22.05 \\
					2.5 & 29.38 \\
					3.0 & 36.01 \\
					\hline
	\end{tabular}
\end{table}

\subsection{In-Bore Experiments}
\begin{figure}[!ht]
	\centering
	\includegraphics[width=0.9\columnwidth]{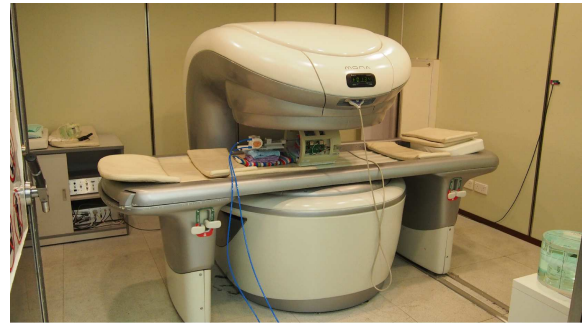}
	\caption{Needle driving mechanism inside the 0.2 T scanner\textsuperscript{\textregistered}}
	\label{0.2T MRI}
\end{figure}

\begin{figure}[!ht]
    \centering
    \includegraphics[width=\columnwidth]{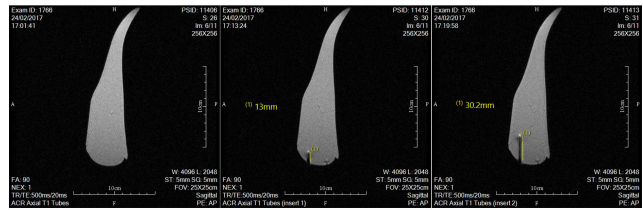}
    \caption{Needle insertion under continuous MRI}
    \label{Insertion_test}
\end{figure}

The magnetic properties and compatibility of the actuator was tested with MRI scanners from Time Medical Systems\textsuperscript{\textregistered}.
We first evaluated the performance of the one degree-of-freedom needle driving mechanism using a $0.2$ T Mona scanner, see Figure \ref{0.2T MRI}.
In this experiment, the motor drives the coaxial needle into a silicon phantom tissue while performing continuous MR imaging. 
Figure \ref{Insertion_test} (left) shows the MR image before the needle was introduced into the phantom; Figure \ref{Insertion_test} (middle) and (right) show the MR image when the needle is inserted at a distance of $13$ mm and $30.2$ mm, respectively, into silicon tissue. 
These results show that the operation of the Tesla motor did not induce any image artifacts during continuous imaging.

\begin{figure}
\centering
\includegraphics[width=0.8\columnwidth]{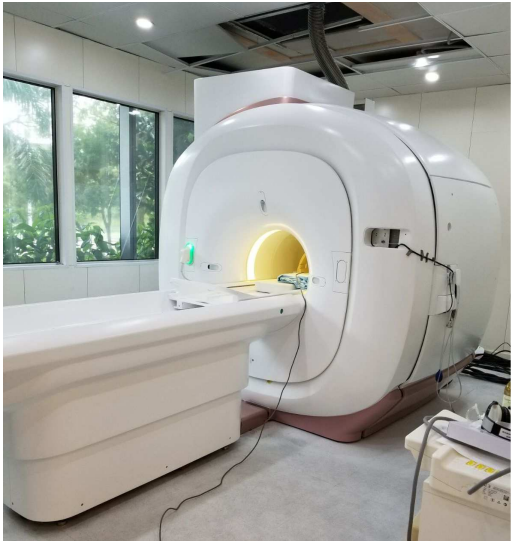}
\caption{A 1-DOF linear joint inside the 1.5T MRI\textsuperscript{\textregistered}}
\label{1.5T MRI}
\end{figure}

The following experiments were all conducted with a high intensity system; we used a $1.5$ T Venus\textsuperscript{\textregistered} scanner from Time Medical, see Figure \ref{1.5T MRI}.
In these experiments, the Tesla motor was placed at the centre of the scanner's magnetic bore; the motor was operated while performing continuous MR imaging.
We evaluated the motor's magnetic compatibility by comparing the computed MR images, signal-to-noise ratio (SNR), homogeneity, and percent integral uniformity (PIU).
These comparison values were obtained from three cases: (1) with only the phantom tissue placed inside the scanner; (2) with the Tesla motor and the phantom placed inside the scanner; (3) with a piezoelectric motor (model GTUSM60, from Glittering Orient Ultrasonic Motor Co\textsuperscript{\textregistered}) and the phantom placed inside the scanner.
Image subtraction was performed to detect geometric changes between the images obtained from cases (1) and (2), and between the images from cases (1) and (3).

SNR can be used to show whether the imaging process is affected by the operation of the actuator or its mechanical structure. 
We computed the SNR for the above-mentioned three cases: the SNR obtained with only the phantom inside the scanner was $454.0$, with both Tesla motor and phantom inside the scanner was $440.0$, and with the piezo-motor and phantom inside the scanner was $409.82$. 
These data shows that the SNR drops $3$\% by operating the Tesla actuator, whereas the piezo-motor makes it drop by almost $10$\%.
Magnetic homogeneity (a value describing the uniformity of the magnetic field) was also computed to evaluate the actuator.
This value is important as it directly related to the image quality and geometric artifacts. 
For this value, the further the robot is placed away from the magnetic field centre, the less it affects the field strength. 
In this experiment, both the Tesla motor and piezoelectric motor were placed at the centre of the field centre, in order to compare their properties. 
The computed homogeneity value with only the phantom placed inside the scanner was $5.56$, with the Tesla motor and phantom was $5.433$, and with the piezo-motor and phantom was $12.982$. 
By comparing these results, it can be seen that the homogeneity value is not considerably affected by the operation of the Tesla motor within the scanner. 
However, the results show that the homogeneity value increases more than double after the piezoelectric motor is introduced to the scanner. 

\begin{figure}
\centering
\includegraphics[width=\columnwidth]{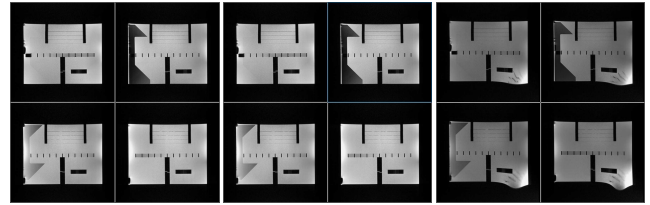}
\caption{MRI scans of a calibration device placed along the: (left) phantom tissue, (middle) phantom and Tesla motor, (right) phantom and piezoelectric motor.}
\label{Sagittal}
\end{figure}

\begin{figure}
\centering
\includegraphics[width=\columnwidth]{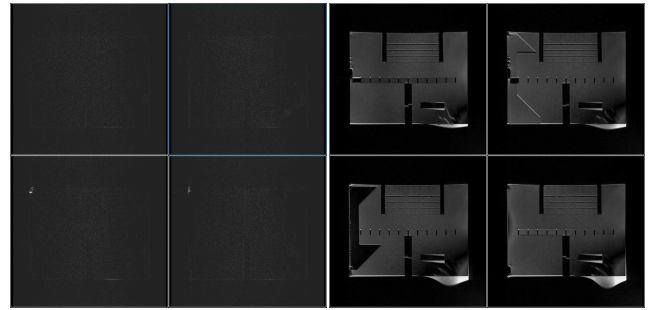}
\caption{Subtraction images of (left) Tesla motor, and (right) piezoelectric motor}
\label{Subtraction}
\end{figure}

The PIU values are useful to quantify geometric distortions of the MR image..
It is the percentage difference between the high and low signal values of the water-only regions in the phantom. 
This PIU value was computed for the three cases under consideration, however, no significant difference was detected amongst the cases.
Finally, we performed a qualitative evaluation of the induced geometric distortions.
This test was conducted by visually inspecting the straightness and lengths of a test calibration device.
Figure \ref{Sagittal} (left) shows the computed MR scan with the phantom only.
Figure \ref{Sagittal} (middle) and (right) show the same phantom scans obtained with the Tesla motor and the piezoelectric motor, respectively.
By comparing these figures, it can be clearly seen that the presence and operation of our Tesla motor does not cause geometric distortions to the image.
The relative distortions caused by these two actuators can be appreciated by inspecting the subtraction images shown in Figure \ref{Subtraction}.
These results prove the feasibility of using our pneumatic rotary actuator to conduct MRI-guided interventions with continuous imaging.

\section{Conclusions}\label{sec:conclusions}
In this paper, we present a new pneumatic rotary motor that can be used as a robot actuator for continuous MR imaging. 
The proposed mechanism is inspired and further modified based on the Tesla turbine: it consists of a blade-less turbine system that exploits the boundary layer effect of the fluid to generate motion (this mechanism enables the generation of continuous rotations in a controllable manner).
To measure the motor's position, we developed a new rotary sensor that counts the pulses of an encoder using optical fibre.
Several experiments were conducted to evaluate the mechanical, control, and magnetic compatibility properties of the system.
The obtained results show that our new actuator can be operated during continuous MRI scans without affecting the scanner.

The motor's design philosophy is to have all the required electronic devices and signals outside the scanning room to prevent interference with MR imaging process.
Since the motor is driven by pneumatic power and its angular position measured with fibre optics, only light and air signals need to be passed to the MRI room through the waveguides (this feature effectively eliminates crucial possible sources of noise and image artifacts).
Compared to other existing pneumatic motors for MRI, the proposed system does not rely on discrete stepping motions; due to its particular mechanical properties, this actuator has the potential to be used as a type of ``servo-motor'' for MRI.
As future work, we would like to improve the dynamic performance of the actuator. Note that for the current system, we only use a simple PID controller for regulating the position via flow control (this method is susceptible to large overshoot for large targets).
Also, we are currently planning the development of a 3-DOF interventional robot that is completely driven by our new Tesla motor.

\begin{table}
				\centering
				\caption{Summary of the compatibility experiments with the $1.5$ T scanner}
				\label{MRI compatibility test}
				\begin{tabular}{p{2.2cm}|p{1.5cm}p{1.5cm}p{1.5cm}}
								\toprule
								\toprule
								MRI Indices & Phantom only & Tesla motor \& phantom & Piezo \& phantom \\
								\midrule
								SNR & 454.0 & 440.0 & 409.82\\
								PIU (\%) & 84.88 & 85.46 & 82.82 \\
								Homogeneity & 5.56 & 5.433 & 12.982 \\
								Geom. distortion &  & Pass & Fail \\
								\bottomrule
				\end{tabular}
\end{table}


%
%
\bibliography{myRef.bib}
\bibliographystyle{IEEETran}

\end{document}